\documentclass{article}
\usepackage{spconf,amsmath,graphicx}
\usepackage{multirow}
\usepackage{subcaption}
\usepackage{color, colortbl}

\definecolor{Gray}{gray}{0.8}

\title{Toward estimating personal well-being using voice}
\name{Samuel Kim, Namhee Kwon, Henry O'Connell}
\address{Canary Speech, LLC, Provo, Utah, U.S.A. \\ \texttt{\small \{sam, namhee, henry\}@canaryspeech.com}}

\begin{document}
%
\maketitle
\begin{abstract}
Estimating personal well-being draws increasing attention particularly from healthcare and pharmaceutical industries. We propose an approach to estimate personal well-being in terms of various measurements such as anxiety, sleep quality and mood using voice. With clinically validated questionnaires to score those measurements in a self-assessed way, we extract salient features from voice and train regression models with deep neural networks. Experiments with the collected database of 219 subjects show promising results in predicting the well-being related measurements; concordance correlation coefficients (CCC) between self-assessed scores and predicted scores are 0.41 for anxiety, 0.44 for sleep quality and 0.38 for mood.

\end{abstract}
\begin{keywords}
Affective computing, well-being, axiety, sleep quality, mood\end{keywords}
\section{Introduction}
\label{sec:intro}

There is a growing body of interest to use voice as one of biomarkers to estimate one's health conditions.  Some health conditions like Parkinson's disease~\cite{Smith2018, Hauptman2019} and amyotrophic lateral sclerosis (ALS)~\cite{Rowe2019} are directly related to speech production mechanism and
others like schizophrenia~\cite{Martinez-Sanchez2015, Voleti2019}, depression~\cite{Dresvyanskiy2019, Espy-Wilson2019}, and bipolar disorder~\cite{Aldeneh2019, Matton2019} are indirectly related through neurological processes that can modulate voice.

This work focuses on personal well-being (or, quality of life) in general population with relatively healthy conditions. Previous related researches include speech emotion recognition (SER)~\cite{Busso2008, 2013ACTI2919} targeting to classify emotional status with a given voice segment. This work expands the scope to the personal well-being; in particular, we study various measurements such as anxiety, sleep quality and mood. 

Like other related applications mentioned above, we also assume that persons' condition is somehow embedded in their voice by modulating articulatory organs either voluntarily or involuntarily. Based on this assumption, we propose an approach to extract salient features from voice with respect to the target measurements and train them with machine learning algorithms to automatically estimate using voice.

In this work, we use subjects' self-assessed measurements as ground truth of their well-being status. Dealing with subjective matters like well-being, however, is difficult partially because it is related to personal perspectives, feelings, and opinions which may vary across the subjects~\cite{citeulike:11036749}. Therefore, we use a questionnaire based approach~\cite{citeulike:11036749, Theodore} to consolidate responses from multiple questions that reflect various aspects of target measurements. 

Details about the questionnaires will be discussed in the next session where we describe the data collection campaign in Section~\ref{secion:data_collection} followed by the proposed method in Section~\ref{section:methodology} and the experimental results in Section~\ref{section:results}.

\section{Data Collection}
\label{secion:data_collection}
\subsection{Procedures}
We recruited 219 participants from Canada for data collection. Besides 13 participants that are not specified their demographic information, there are 114 males and 92 female and their age range is from 25 to 55 (33.6 $\pm$ 8.1) years. Their native language is English including 12 French bilingual subjects. They were also asked to report their medical conditions. Being allowed multiple choices, no previous medical condition was reported for 159 participants while some reported that they have been treated due to some medical conditions notably 22 with depression and 13 with migraine or headache. 

The data collection was designed to conduct five sessions per participants with several days of intervals in between. During the data collection campaign, 202 participants completed all five sessions and the intervals between sessions per participants are from 1.2 to 15.3 (3.7 $\pm$ 1.9) days. At each session, the participants were asked to answer a set of questionnaires using mobile devices; four written surveys and a questionnaire that requires voice responses. 

After removing sessions that are incomplete or have missing fields, we have total 1,048 sessions collected (approx. 4.8 sessions per subjects).

\subsection{Questionnaires}
\label{subsecion:questionnaires}

\subsubsection{Written surveys}
As discussed earlier, we use a questionnaire based approach to collect self-assessed measurements to represent subjects' well-being status. In this regard, we adopt clinically validated questionnaires that are designed to measure anxiety, sleep quality and mood as follows. 

\begin{itemize}
    \item Anxiety: State-trait anxiety inventory (STAI) \cite{Tluczek2009} and Generalized anxiety disorder (GAD7) \cite{10.1001/archinte.166.10.1092} are used to measure the level of anxiety. STAI has six emotional status to score themselves based on how they feel at the moment, while GAD7 has seven emotional status to score how often they felt over the past two weeks. Both have consolidating rules to generate the level of anxiety based on the answers; higher value means higher anxiety. 
    \item Sleep quality: We use Pittsburgh sleep quality index (PSQI) \cite{BUYSSE1989193} which was design to measure sleep quality. The questionnaire comprises various aspects of sleep quality such as length of sleep as well as disturbing factors and their frequencies. The scoring guideline is to generate the level of sleep discomfort; higher value indicates worse sleep quality. 
    \item Mood: We use positive affect and negative affect schedule (PANAS) \cite{Watson1988}. There are ten different emotional status (five positive and five negative) to answer what extent the subject had felt over the past week. It gives an aggregated score to represent status of the subject; higher value indicates more negative mood and lower value indicates more positive mood. 
\end{itemize}

Table~\ref{table:measurements} summaries the statistics of collected scores of the measurements along with possible score ranges. Note that the collected data covers possible range of individual measurements except PSQI covers only lower range.

\subsubsection{Voice responses}
The questionnaire that requires voice responses is designed to capture vocal behaviors and to use them in estimating the above described measurements. We asked seven different questions to elicit three types of voice responses; one spontaneous speech, four sentence readings and two paragraph readings. 

For spontaneous speech, participants were given an instruction to speak freely on whatever topic they want for about a minute. For sentence or paragraph readings, phonetically balanced reading materials are prompted to the subjects so that they can read aloud (8.5 words for sentence readings and approx. 130 words for paragraph readings on average). 

Table~\ref{table:audio_length} shows the statistics of collected voice responses in terms of recorded length. Overall, we collected approximately 54 hours of voice data. 

    

\begin{table}[t]
  \caption{Statistics of self-assessed measurements.}
  \label{table:measurements}
  \centering
  \begin{tabular}{c||c|c|c}
  \hline
  \multirow{2}{*}{} & \multirow{2}{*}{Possible range} & \multicolumn{2}{c}{Collected Data} \\ \cline{3-4}
  & & Range & Mean $\pm$ STD\\
  \hline \hline
  STAI & 20 - 80 & 20 - 80 & 38.4 $\pm$ 13.2  \\ \hline
  GAD7 & 0 - 21 & 0 - 21 & 6.1 $\pm$ 4.9 \\ \hline
  PSQI & 0 - 21 & 0 - 15 & 5.3 $\pm$ 2.8  \\ \hline
  PANAS & 10 - 50 & 10 - 50 & 24.4 $\pm$ 6.7~~  \\ \hline
  \end{tabular}
\end{table}


\begin{table}[t]
  \caption{Length of voice responses.}
  \label{table:audio_length}
  \centering
  \begin{tabular}{c|c||c}
  \hline
  \multicolumn{2}{c||}{Voice responses} & Length (in secs.) \\
   \hline \hline
  Spontaneous & Q1 & 64.0 $\pm$ 10.3\\ \hline
  \multirow{4}{*}{Sentence reading} & Q2 &  5.4 $\pm$ 1.5 \\ \cline{2-3}
   & Q3 &  5.1 $\pm$ 1.6 \\ \cline{2-3}
   & Q4 &  4.8 $\pm$ 1.6 \\ \cline{2-3}
   & Q5 &  5.1 $\pm$ 1.6 \\ \hline
  \multirow{2}{*}{Paragraph reading} & Q6 & 48.6 $\pm$ 8.6~~ \\ \cline{2-3}
    & Q7 & 48.6 $\pm$ 12.4 \\ \hline
  \end{tabular}
\end{table}

\begin{table*}[t]
  \caption{Concordance correlation coefficients (CCC) between self-assessed scores and predicted scores using voice responses. Statistical significance is denoted with stars ($^{*}$ for $p<10^{-2}$ and $^{**}$ for  $p<10^{-5}$). }
  \label{table:per_individual_questions}
  \centering
  \begin{tabular}{c|c||>{\hspace{1.2pc}}l|l|l|l|l|l|l||>{\hspace{1.4pc}}l}
  \hline
  \multicolumn{2}{c||}{} & \multicolumn{7}{c||}{Individual features} & \multicolumn{1}{c}{\multirow{3}{*}{\shortstack[c]{Concatenated \\ features}}}  \\
  \cline{3-9}
   \multicolumn{2}{c||}{} & \multicolumn{1}{c|}{Spontaneous} & \multicolumn{4}{c|}{Sentence reading} &
   \multicolumn{2}{c||}{Paragraph reading}  & \\
   \cline{3-9}
   \multicolumn{2}{c||}{} & \multicolumn{1}{c|}{Q1} & \multicolumn{1}{c|}{Q2} &\multicolumn{1}{c|}{Q3} &\multicolumn{1}{c|}{Q4} &\multicolumn{1}{c|}{Q5} &\multicolumn{1}{c|}{Q6} &\multicolumn{1}{c||}{Q7} &  \\
    \hline \hline
  \multirow{2}{*}{STAI} & eGeMAPS &  0.05 & 0.03 & 0.07$^{*}$ & 0.02 & 0.04 & 0.03 & 0.07 & 0.04$^{**}$ \\ \cline{2-10}
& \cellcolor{Gray} Proposed &\cellcolor{Gray} 0.09$^{**}$ &\cellcolor{Gray} 0.16$^{**}$ & \cellcolor{Gray}0.19$^{**}$ &\cellcolor{Gray} 0.15$^{**}$ & \cellcolor{Gray}0.14$^{**}$ &\cellcolor{Gray} 0.14$^{**}$ & \cellcolor{Gray}0.14$^{**}$ &\cellcolor{Gray} 0.30$^{**}$ \\ \hline
\multirow{2}{*}{GAD7} & eGeMAPS &0.02 & 0.05$^{*}$ & 0.04 & 0.03 & 0.01 & 0.05$^{*}$ & 0.04$^{*}$ & 0.03$^{**}$\\ \cline{2-10}
 & \cellcolor{Gray} Proposed &\cellcolor{Gray}0.14$^{**}$ &\cellcolor{Gray} 0.19$^{**}$ &\cellcolor{Gray} 0.26$^{**}$ &\cellcolor{Gray} 0.22$^{**}$ & \cellcolor{Gray}0.17$^{**}$ &\cellcolor{Gray} 0.08$^{**}$ &\cellcolor{Gray} 0.09$^{**}$ &\cellcolor{Gray} 0.41$^{**}$ \\ \hline
\multirow{2}{*}{PSQI} & eGeMAPS &0.03 & 0.08$^{*}$ & 0.04 & 0.01 & 0.02 & 0.10$^{**}$ & 0.10$^{**}$ & 0.06$^{**}$\\ \cline{2-10}
&  \cellcolor{Gray}Proposed &\cellcolor{Gray}0.14$^{**}$ & \cellcolor{Gray}0.17$^{**}$ & \cellcolor{Gray}0.21$^{**}$ & \cellcolor{Gray}0.20$^{**}$ &  \cellcolor{Gray}0.21$^{**}$ & \cellcolor{Gray}0.12$^{**}$ &  \cellcolor{Gray}0.09$^{**}$ &\cellcolor{Gray}0.44$^{**}$ \\ \hline
\multirow{2}{*}{PANAS} & eGeMAPS &-0.01 & 0.00 & 0.01 & 0.01 & 0.00 & 0.03 & 0.05$^{*}$ & 0.02$^{*}$\\ \cline{2-9}
& \cellcolor{Gray}Proposed &\cellcolor{Gray}0.12$^{**}$ & \cellcolor{Gray}0.18$^{**}$ & \cellcolor{Gray}0.20$^{**}$ & \cellcolor{Gray}0.16$^{**}$ & \cellcolor{Gray}0.17$^{**}$ & \cellcolor{Gray}0.12$^{**}$ &\cellcolor{Gray} 0.15$^{**}$ & \cellcolor{Gray}0.38$^{**}$\\ \hline
  \end{tabular}
\end{table*}

\section{Methodology}
\label{section:methodology}
\subsection{Feature extraction and selection}
The types of features are in two categories, i.e. acoustic and linguistic features. Acoustic features are to capture signal-level modulations due to speakers' status, while linguistic features are to capture language-level patterns that may be influenced by the condition. 


Acoustic features are calculated on a per-frame basis. Frames are defined as 25 ms sliding windows that are created every 10 ms. A 41-dimensional supervector of various features such as mel-frequency cepstral coefficients (MFCC), perceptual linear prediction (PLP), prosody and voice quality related features is generated every frame, and its delta and delta-delta are concatenated to capture frame-level context. To summarize, we use 19 statistical functions such as mean, median, skewness, kurtosis, quartile, percentile, slope, etc to generate a response-level feature vector.

Language features are based on the results from automatic speech recognition (ASR). We used Canary's general English model which is trained on publicly available datasets like Tedlium and Librispeech using the time delayed neual network (TDNN) architecture in Kaldi~\cite{Povey_ASRU2011}. On top of common features such as part-of-voice ratio, syllable duration, filler ratio and word repetition ratio over the total number of spoken words, we extract a different feature set whether the response is spontaneous or read. 

For the spontaneous voice responses where no prompted text was given, the semantic features are extracted including word popularity percentile\footnote{The word popularity was computed from the general English language model with 130K words and the distribution of the values per response was examined and its percentile values were used.}, and word frequency of the depression-related terms\footnote{We built a dictionary with depression-related words and negative expressions, as well as common words observed from depression patients' speech, which resulted in 486 terms~\cite{doi:10.1177/2167702617747074}
}, and positive and negative sentiment likelihood\footnote{The sentiment likelihood score is generated by the binary classification model trained on Stanford Sentiment Treebank using Sentiment Neuron~\cite{RadfordJS17}.}.  For the read responses, on the other hand, the ASR errors for insertion, deletion, and substitution are computed based on the given text. 

The feature dimension is 2,357 for a read response and 2,364 for a spontaneous response. For feature selection, we compute the Pearson's correlation coefficient between the extracted features and individual self-assessed measurement scores and selected top~\textit{n} correlated features for the model. Note that it is done in response- and measurement- level so that different features can be selected from different responses depending on the target measurement.

\begin{figure*}[t!]
  \centering
  \begin{subfigure}{0.4\textwidth}
	  \includegraphics[width=\linewidth]{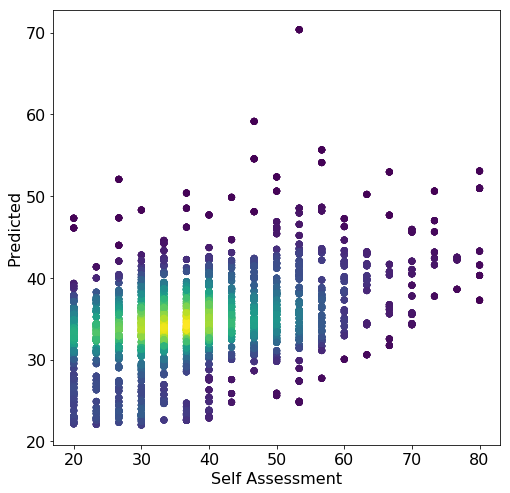}
	  \caption{STAI}
  \end{subfigure}
  \begin{subfigure}{0.4\textwidth}
	  \includegraphics[width=\linewidth]{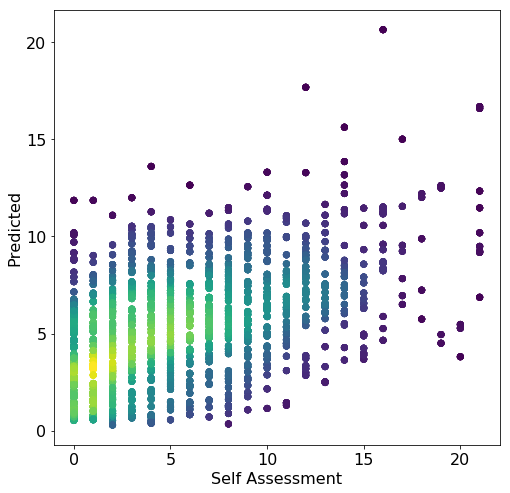}
	  \caption{GAD7}
  \end{subfigure}
  \begin{subfigure}{0.4\textwidth}
	  \includegraphics[width=\linewidth]{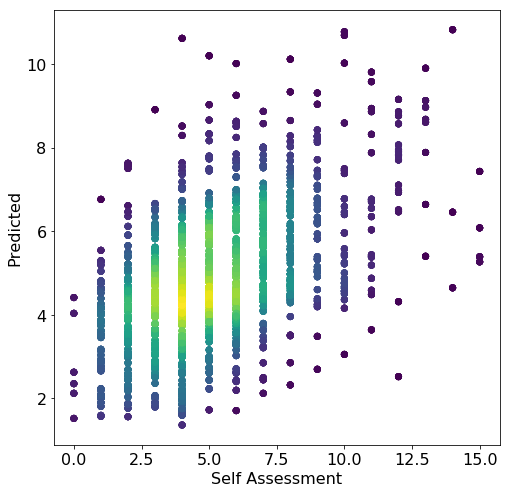}
	  \caption{PSQI}
  \end{subfigure}
  \begin{subfigure}{0.4\textwidth}
	  \includegraphics[width=\linewidth]{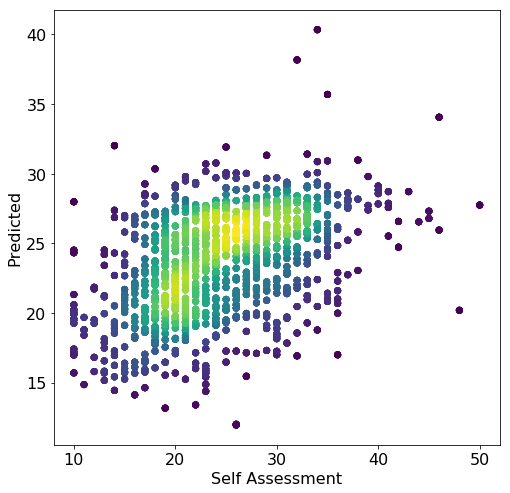}
	  \caption{PANAS}
  \end{subfigure}
  \caption{Scatter plots of self-assessed scores and predicted scores using the proposed method. Colors represent the density of the individual data points (yellow for high density and blue for low density).}
        \label{fig:scatter}
\end{figure*}




As a baseline, we use OpenSmile toolkit~\cite{Eyben:2013:RDO:2502081.2502224} with eGeMAPS configuration~\cite{Eyben:2015} which is widely used and shows promising results particularly in affective computing related fields such as speech emotion recognition. As it generates a 88 dimensional feature vector per one speech signal, we set a threshold of the proposed feature selection procedure to retain only the same number of features for fair comparisons. 

\subsection{Modeling}
We use fully connected deep neural network (FC-DNN) with 4 hidden layers of 256 neuron units. Each layer has the rectified linear unit (ReLU) as an activation function, and l2 regularizer and 50\% of dropout to prevent overfitting. The output layer has only one unit as we target to build a regression model. We used an Adam optimizer with learning rate 0.0001 using mean squared error (MSE) as a loss function. The batch size is 32 and iteration stops after 100 epochs.

\section{Experimental Results}
\label{section:results}
\subsection{Setup}

We perform a 5-fold cross validation; we randomly split the whole dataset into 5 exclusive folds and iteratively use one fold as test data and others as training data. Each fold is subject independent in a sense that different folds do not share data from the same subject. Note that we consider all the sessions independent. A longitudinal study to investigate changes over time is left for future work.

The performance is measured in terms of concordance correlation coefficients (CCC) between self-assessed scores and predicted scores~\cite{10.2307/2532051}. 

\subsection{Results}
Table~\ref{table:per_individual_questions} shows CCC between self-assessed scores and predicted scores using voice responses with respect to the target measurement. The left part of the table shows the results with features from individual voice responses, while the right part of the table shows the results with concatenated features from individual voice responses. Fig.~\ref{fig:scatter} shows the scattered plot of self-assessed measures and estimated values using concatenated features. 


The results with individual voice responses show that the proposed method outperforms the conventional general purpose feature extraction method in estimating all measurements with all voice responses. The same applies to the results with concatenated features. This indicates that the proposed feature selection strategy that considers differences in individual responses is beneficial. The proposed linguistic features are also considered to contribute toward the improvement. 

It is notable that the correlation coefficients with features from sentence reading are higher than the ones with features from spontaneous or paragraph reading responses. The results also show that concatenating features from individual voice responses boosts the performance. These suggest that concatenating features from multiple short voice responses rather than one long voice response with a fixed set of features is beneficial to estimate self-assessed measurements. It may be partially because that salient features are averaged out over a long sentence, but in-depth study will be in the future. In the future, we will also study impact of the ASR performance in extracting linguistic features as well as longitudinal study to investigate changes over time.

\section{Conclusion}
We showed promising results in estimating well-being status with voice in terms of various measurements, i.e. anxiety, sleep quality, and mood. The proposed method utilizes acoustic and linguistic features along with response- and measurement- level feature selection strategy followed by a deep neural network based regression model. Experimental results show that sleep quality using PSQI has the strongest correlation while anxiety using STAI has the weakest correlation. Although the concordance correlation coefficients between self-assessed scores and estimated scores may not be too strong, they are statistically significant with $p<10^{-5}$.

\section{Acknowledgement}
\label{sec:ack}
The authors like to thank Nathan Fisk, Scott Ferguson and Mark Bartlett for valuable discussions.

\vfill\pagebreak
\newpage

\bibliographystyle{IEEEbib}
{\small \bibliography{mybib}}

\end{document}